\documentclass[a4paper,fleqn]{cas-dc}

\usepackage[numbers,sort&compress,longnamesfirst]{natbib}
\usepackage{subfigure}
\usepackage{amssymb}
\usepackage{gensymb}

\usepackage{color, xcolor}
\usepackage{hyperref}

\def\tsc#1{\csdef{#1}{\textsc{\lowercase{#1}}\xspace}}
\tsc{WGM}
\tsc{QE}

\tsc{EP}
\tsc{PMS}
\tsc{BEC}
\tsc{DE}
\begin{document}
\let\WriteBookmarks\relax
\def\floatpagepagefraction{1}
\def\textpagefraction{.001}

\setlength\abovedisplayskip{1pt}
\setlength\belowdisplayskip{1pt}

% Short title
\shorttitle{MLP-AIR: An Efficient MLP-Based Method for Actor Interaction Relation Learning in Group Activity Recognition}    
% Short author
\shortauthors{Guoliang Xu et~al.}  

% Main title of the paper
\title [mode = title]{MLP-AIR: An Efficient MLP-Based Method for Actor Interaction Relation Learning in Group Activity Recognition}  

\author[]{Guoliang Xu}[type=editor,
                        auid=000,
                        bioid=1,
%                        prefix=Sir,
%                        role=Researcher,
%                        orcid=0000-0001-7511-2910
]

\ead{xgl@bupt.edu.cn}
%\ead[url]{www.cvr.cc, cvr@sayahna.org}
%\credit{Conceptualization of this study, Methodology, Software}
%\address[1]{Elsevier B.V., Radarweg 29, 1043 NX Amsterdam, The Netherlands}
\address{School of Artificial Intelligence, Beijing University of Posts and Telecommunications, Beijing, China.}

\author[]{Jianqin Yin}[%
%   role=Co-ordinator,
%   suffix=Jr,
   ]
\cormark[1]
%\fnmark[2]

\ead{jqyin@bupt.edu.cn}
%\ead[URL]{www.sayahna.org}

\cortext[cor1]{Corresponding author}

% Here goes the abstract
\begin{abstract}
The task of Group Activity Recognition (GAR) aims to predict the activity category of the group by learning the actor spatial-temporal interaction relation in the group. Therefore, an effective actor relation learning method is crucial for the GAR task. The previous works mainly learn the interaction relation by the well-designed GCNs or Transformers. For example, to infer the actor interaction relation, GCNs need a learnable adjacency, and Transformers need to calculate the self-attention. Although the above methods can model the interaction relation effectively, they also increase the complexity of the model (the number of parameters and computations). In this paper, we design a novel MLP-based method for Actor Interaction Relation learning (MLP-AIR) in GAR. Compared with GCNs and Transformers, our method has a competitive but conceptually and technically simple alternative, significantly reducing the complexity. Specifically, MLP-AIR includes three sub-modules: MLP-based Spatial relation modeling module (MLP-S), MLP-based Temporal relation modeling module (MLP-T), and MLP-based Relation refining module (MLP-R). MLP-S is used to model the spatial relation between different actors in each frame. MLP-T is used to model the temporal relation between different frames for each actor. MLP-R is used further to refine the relation between different dimensions of relation features to improve the feature’s expression ability. To evaluate the MLP-AIR, we conduct extensive experiments on two widely used benchmarks, including the Volleyball and Collective Activity datasets. Experimental results demonstrate that MLP-AIR can get competitive results but with low complexity.
\end{abstract}

% Use if graphical abstract is present
%\begin{graphicalabstract}
%\includegraphics{}
%\end{graphicalabstract}

% Research highlights

% Keywords
% Each keyword is seperated by \sep

\maketitle
% Main text
\section{Introduction}

Like action recognition, GAR is an important problem in video understanding and has been widely applied, such as surveillance, sports video analysis, and social behavior understanding\cite{1_Deng2022Summarization,2_ehsanpour2020joint}. Unlike action recognition, GAR aims to predict the activity of the group\cite{3_liu2022visual,4_LU2018195}, while action recognition aims to predict the action of individuals\cite{5_WANG2021265,6_ZHENG202210,7_WU2022358,8_YUE2022287}. Therefore, actor interaction relation learning is more critical in GAR\cite{9_PEREZ2022108360,10_wu2019learning,11_gavrilyuk2020actor,12_zhou2021composer}.

Recently, deep learning methods have achieved significant results in GAR. The deep learning methods mainly use well-designed GCNs\cite{10_wu2019learning,13_9241410,14_yan2020social,15_yuan2021spatio} and Transformers\cite{11_gavrilyuk2020actor,12_zhou2021composer,16_han2022dual,17_li2021groupformer,18_li2022learning} to model the actor interaction relation. In GCN-based methods, these methods aim to model the pair-wise actor interaction relation with an adjacency matrix, where the graph nodes represent the feature of actors, and the graph edges represent the relation of different nodes. Typically, the adjacency matrix can be optimized in an end-to-end manner. In transformer-based methods, these methods model the interaction relation between different actors by the self-attention mechanism and select the critical information for GAR, where each actor feature is viewed as embedding. We find that the previous methods improve the performance but also the complexity.

\begin{figure*}[Figure 1]
    \centering
    \subfigure{
        \includegraphics[width=1\textwidth]{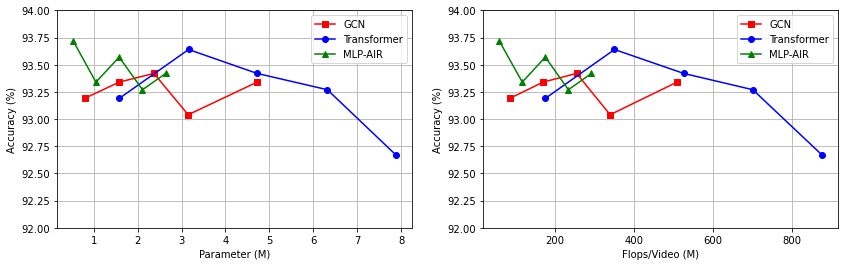}
    }
\begin{flushleft}
\caption{Our MLP-AIR vs. GCN and Transformer in the unified framework for GAR. Left: the number of parameters; Right: the number of computations.}\label{Figure 1}
\end{flushleft}
\end{figure*}

Hence, we propose an MLP-based method for actor interaction learning to reduce the complexity of relation methods. Compared with GCNs and Transformers, MLP does not need the complex convolution and self-attention mechanism but is based entirely on the multi-layer perceptron, which provides a conceptually and technically simple alternative\cite{19_zhao2021battle}. \cite{20_tolstikhin2021mlp,21_zhang2022morphmlp,22_chen2021cyclemlp} successively propose MLP-based backbones to extract features of images and videos for various tasks, such as image classification and video classification, etc. Without the help of convolution and self-attention mechanisms, they also can effectively extract features using MLP. To our best knowledge, we are the first to use MLP to model the actor interaction relation. Unlike them, we use MLP for actor interaction relation learning instead of image/video representation extraction. As shown in Fig. \ref{Figure 1}, we evaluate the performance of GCN, Transformer, and MLP fairly under a unified framework. We can find that MLP can get the best performance with lower complexity than other methods.

It is difficult for MLP to model the spatial-temporal relation. From the relation modeling perspective, GCN and Transformer use the adjacency matrix and attention matrix to model the spatial-temporal interaction between actors explicitly. However, MLP does not contain a similar matrix describing relations. It only uses the multi-layer perceptron to model these relations. Therefore, it is worth considering how to design the fully connected layer for spatial-temporal interaction relation learning.

\begin{figure*}[Figure 2]
    \centering
    \subfigure{
        \includegraphics[width=1\textwidth]{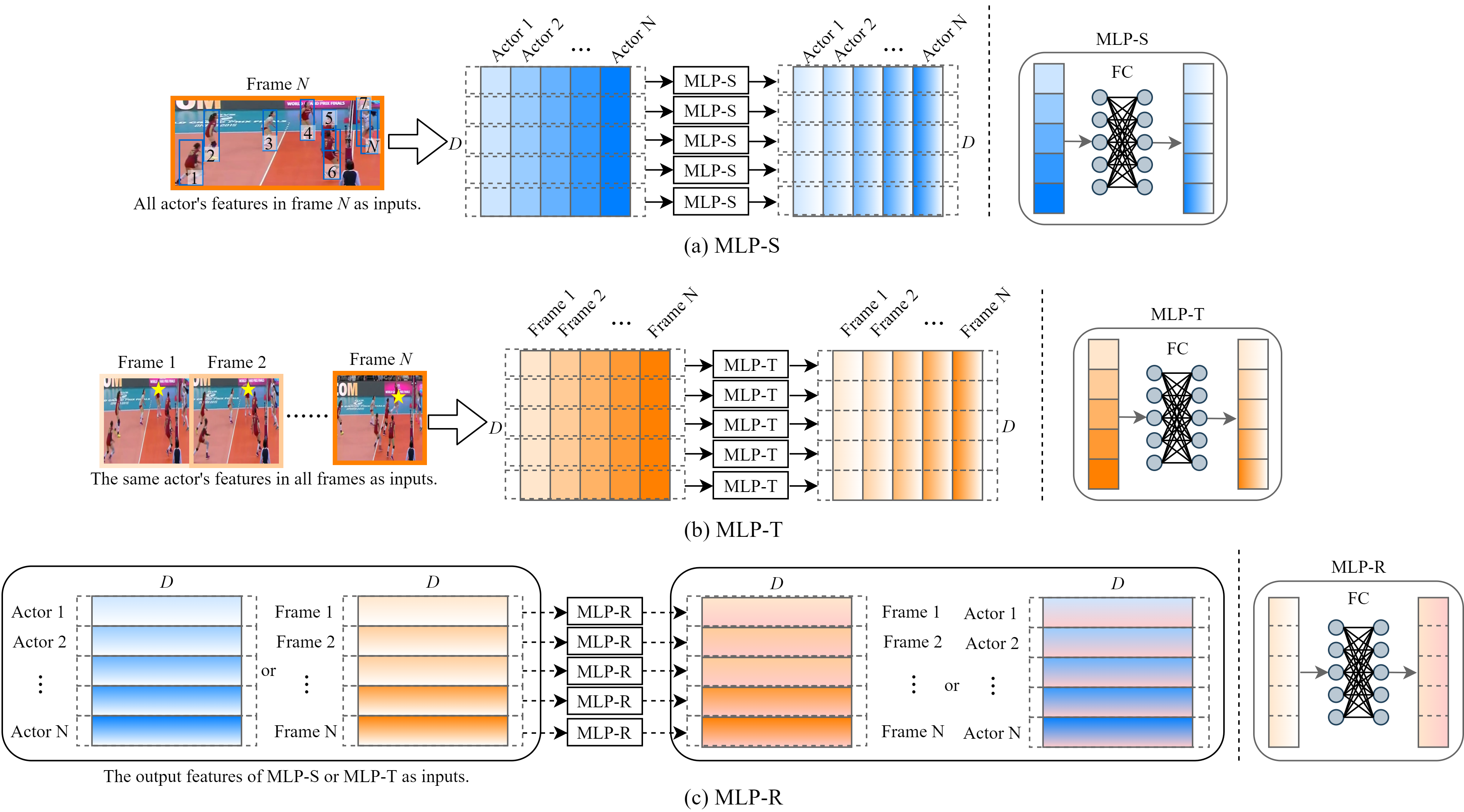}
    }
    
\begin{flushleft}
\caption{The overview of three sub-modules. The shape of each feature is 1×D. (a) MLP-S models the spatial interaction relation of all actors within the same frame; (b) MLP-T models the temporal interaction relation of the same actors in all frames. The same actors in all frames are marked by pentagrams; (c) MLP-R further refines the relation between different dimensions of relation features.
}\label{Figure 2}
\end{flushleft}
\end{figure*}

To tackle this challenge, we propose the MLP-based method to learn the actor spatial-temporal relation, namely MLP-AIR. Unlike GCNs and Transformers, to achieve spatial-temporal relation modeling, the key problem is that MLP-AIR needs to achieve across-actor or across-time relation modeling implicitly. As shown in Fig. \ref{Figure 2}(a), we use MLP to learn the features between different actors in each frame to achieve the goal of spatial relation modeling. As shown in Fig. \ref{Figure 2}(b), we also use MLP to learn the feature of the same actor at different times to achieve the goal of temporal relation modeling.

Specifically, MLP-AIR mainly includes three sub-modules: MLP-S, MLP-T, and MLP-R. \textbf{(1) MLP-S} uses the concise FC (Fully Connected layer) to model the spatial interaction relation. As shown in Fig. \ref{Figure 2}(a), the main reason is that MLP-S can realize cross-actor relation learning, which operates in the same dimension of all actor’s features. The feature matrix is composed of the features of all actors in the same frame, and MLP-S takes the rows of the feature matrix as inputs. \textbf{(2) MLP-T} also uses the concise FC to model the temporal interaction relation. As shown in Fig. \ref{Figure 2}(b), the main reason is that MLP-T can realize across-frame relation learning of the same actor, which operates in the same dimension of all frame’s features. The feature matrix is composed of the same actor’s features in all frames, and MLP-T takes the rows of the feature matrix as inputs. We can find that MLP-S and MLP-T only model the relation in the same dimension of each feature without considering the relation of different dimensions. Therefore, we also propose \textbf{(3) MLP-R}, which is skilled in across-dimensions relation modeling, improving the expression ability of relation. We use the FC to refine each feature, as shown in Fig. \ref{Figure 2}(c). The main reason is that MLP-R can realize across-dimension relation learning, which operates in the different dimensions of the same feature. MLP-R takes the rows of the feature matrix (each feature) as inputs.

Then, we combine these basic modules: MLP-S, MLP-T, and MLP-R, according to the different demands. Among them, we build the SRM (Spatial Relation Module) by arranging MLP-S and MLP-R in sequence and stack multiple SRMs to realize a multi-layer spatial interaction relation modeling. In the same way, we build the TRM (Temporal relation Module) by arranging MLP-T and MLP-R in sequence and stack multiple TRMs to realize a multi-layer temporal interaction relation modeling. Finally, inspired by \cite{16_han2022dual}, MLP-AIR also establishes a dual-path actor interaction relation learning framework for GAR according to different spatial-temporal orders. Among them, one path is in the order of space before time, while the other is in the order of time before space.

To our best knowledge, we are the first to propose an MLP-based method for actor interaction relation learning. Compared with the previous methods, MLP-AIR significantly reduces the model’s complexity and with better accuracy. The main contributions of this research are summarized as follows:

(1)	We first propose an MLP-based method for actor interaction relation learning in GAR. Since our method provides a conceptually and technically simple alternative, it can significantly reduce the model’s complexity but with competitive accuracy.

(2)	We have designed MLP-AIR reasonably and proposed MLP-S, MLP-T, and MLP-R to implicitly model the spatial-temporal interaction relation of actors and use MLP-R to improve the expression ability of relation further.

(3)	Experiments on two public benchmarks demonstrate that our proposed method can get competitive accuracy with lower complexity. Furthermore, we further demonstrate the effectiveness of our proposed modules through extensive ablation studies.

\section{Related Work}

Recently, GAR has attracted more attention from scholars. At present, a large number of advanced methods have emerged. We can divide these methods into methods without actor interaction learning \cite{23_borja2020deep,24_azar2018multi,25_tsunoda2017football,26_ramanathan2016detecting} and with actor interaction learning \cite{27_qi2019stagnet,28_yan2018participation}. Since group activity involves many actors, it is necessary to consider the impact of actor interaction relations. Therefore, the methods with interaction learning are becoming increasingly important \cite{10_wu2019learning,11_gavrilyuk2020actor,17_li2021groupformer,27_qi2019stagnet,28_yan2018participation}. These methods aim to analyze the actor spatial-temporal interaction relation delicately for GAR. This is a fine-grained way, and our method belongs to this way. Currently, GCN and Transformer are the main ways to model the actor interaction relation in these methods \cite{10_wu2019learning,11_gavrilyuk2020actor,14_yan2020social,17_li2021groupformer}.

\textbf{GCN based methods:} T. N. Kipf et al. \cite{29_kipf2016semi} propose that GCN is applied to feature learning of graph-structured data and achieves significant results. Then, GCN becomes increasingly popular and is used for other graph-structured data, such as human skeleton data \cite{30_yan2018spatial,31_duan2022revisiting}, point cloud data \cite{32_kim2022low,33_fei2022comprehensive}, etc. Wu et al. \cite{10_wu2019learning} first propose ARG (Actor Relation Graph) by the GCN to model the relation of appearance and position between actors. Specifically, they build an adjacency matrix to explicitly model pair-wise actor interaction relation, where the value of the matrix represents the importance of one actor to another actor. Then, the graph-based actor relation modeling methods have been continuously developed. Yan et al. \cite{13_9241410} propose the cross-inference mechanism that discards many unessential connections between each actor, which simplifies the relation matrix. Yuan et al. \cite{15_yuan2021spatio} propose DR (Dynamic relation module) and DW (Dynamic walk module). DR is responsible for predicting the relation matrix, and DW is responsible for predicting the dynamic walk offsets in a joint-processing manner. The person–specific graph is formed in this way. However, GCN is limited by a predefined graph in GAR, which can directly affect the reasoning of interaction relations. The predefined graph usually is built by the spatial-temporal prior information.

In this paper, we aim to use the MLP-based method for actor interaction relation learning. Compared with GCNs, our method has the following advantages. Firstly, our method does not need to build a relation matrix to describe the actor interaction relation, but use the FC operation to implicitly model the interaction relation, which simplifies the model’s complexity. Secondly, our method considers the global relation of all actors, which is not limited by the predefined graph.

\textbf{Transformer based methods:} A. Vaswani et al. \cite{34_NIPS2017_3f5ee243} first propose that Transformer is used to model the global dependency of the sequence in NLP. Subsequently, a couple of Transformer-based works have made a series of breakthroughs in computer vision task, which make Transformer gradually dominate the main position \cite{35_dosovitskiy2020image,36_wang2021pyramid,37_srinivas2021bottleneck,38_liu2021swin,39_yang2021focal,40_fan2021multiscale}. K. Gavrilyuk et al. \cite{11_gavrilyuk2020actor} first apply Transformer to GAR. They regard different actor features as input sequences and then model the actor interaction relation by the self-attention mechanism. To avoid ignoring the actor's position information, they encode the actor's position as the additional input. Similarly, Han et al. \cite{16_han2022dual} introduce the actor’s temporal encoding into Transformer for GAR. Due to Transformer's excellent global relation modeling ability, a series of works have directly applied Transformer to their work as a relation modeling module. We can find that the Transformer-based methods need the self-attention mechanism and spatial-temporal prior information to achieve the relation modeling.

In this paper, compared with Transformers, we propose a self-attention free method that can model actor interaction relation without the spatial-temporal prior information. In this way, our method ensures competitive accuracy and simplifies the model's complexity.

\textbf{MLP for Feature Extraction:} Recently, research shows that the self-attention mechanism may not be necessary for computer vision. More researchers are using simple MLP instead of self-attention for feature extraction \cite{20_tolstikhin2021mlp,21_zhang2022morphmlp,41_hou2022vision,42_tang2022sparse,43_touvron2022resmlp,44_NEURIPS2021_4cc05b35}. I. Tolstikhin et al. \cite{20_tolstikhin2021mlp} propose MLP-Mixer as the backbone for visual tasks in the image domain. MLP-Mixer includes channel-mixing MLPs and token-mixing MLPs. The former is used to learn the features of different channels, and the latter is used to learn the features of different tokens. Subsequently, D. J. Zhang et al. \cite{21_zhang2022morphmlp} propose MorphMLP as the backbone in the video domain. MorphMLP contains MorphFCs and MorphFCt modules used to learn the video's spatial and temporal features. Compared with GCNs and Transformers, MLP provides a conceptually and technically simple alternative. However, to our best knowledge, MLP has not been explored in GAR to model the actor interaction relation.

In this paper, compared with MLP-Mixer, we have significant differences. MLP-Mixer uses the image patches as inputs to extract the image represents for various downstream tasks. Unlike them, we use the actor's features as inputs to model the actor spatial-temporal interaction relation by MLP for GAR. More importantly, we first introduce MLP into GAR to explore its actor interaction relation learning ability.

\begin{figure*}[Figure 3]
    \centering
    \subfigure{
        \includegraphics[width=1\textwidth]{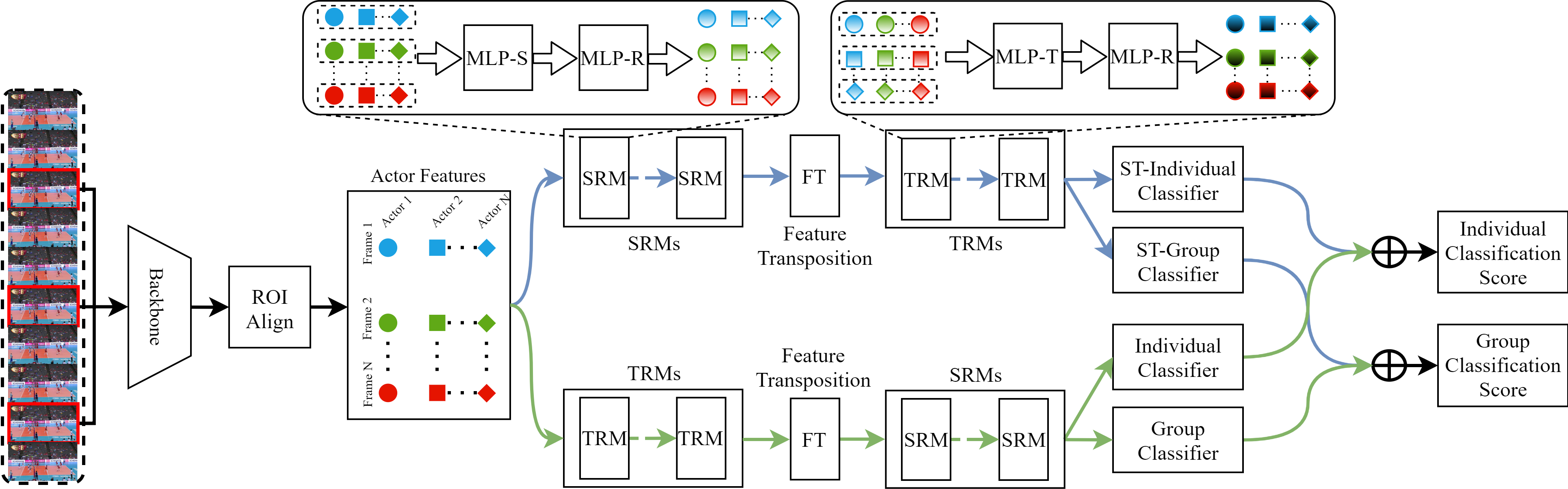}
    }
\begin{flushleft}
\caption{The overall framework of our approach.}\label{Figure 3}
\end{flushleft}
\end{figure*}

\section{Methodology}

The actor's interaction information is crucial for activity recognition in the multi-person scene. Therefore, we propose the MLP-based method to model the actor interaction relation and further obtain the group-level information for GAR. In this section, we will introduce our method in detail. Firstly, we will introduce the overall framework of our approach. Secondly, we will introduce how to realize actor interaction relation learning, including spatial relation learning, temporal relation learning, relation feature refinement, and dual-path actor relation learning. Finally, we will introduce how to optimize our network by loss functions.

\subsection{Overall Framework}

The overall framework of our method is shown in Fig. \ref{Figure 3}. Given the video sequence and actor's bounding boxes, our approach's process can be divided into three steps. In the first step, we randomly select \emph{T} frames from the sequence as the model's inputs and then use the backbone to extract the multi-scale feature map of frames. Like the previous works \cite{10_wu2019learning,16_han2022dual,45_bagautdinov2017social}, we use Inception-v3 \cite{46_szegedy2016rethinking} as the backbone to ensure comparison fairly. Then, we take the feature map as the input of RoIAlign \cite{47_he2017mask} and align the features in the bounding box of each actor. To obtain each actor’s feature, we use 1×1 convolution and full connection layer to map the aligned feature. Then, we use the matrix $ X \in {R^{T \times N \times D}} $ to represent each actor’s feature in all frames, where \emph{N} represents the number of actors in each frame, and \emph{D} denotes the dimension of the actor's feature.

In the second step, after getting each actor’s feature, we next model the actor interaction relation. Inspired by \cite{16_han2022dual}, we also design a dual-path to model the interaction relation of different spatial-temporal orders. But unlike them, we design a novel actor interaction relation learning module by MLP, including MLP-S, MLP-T, and MLP-R. MLP-S and MLP-R constitute the SRM, which is used to model and refine spatial relation features. MLP-T and MLP-R constitute the TRM, which is used to model and refine temporal relation features. We put these modules in a different order so that the two paths (spatial-temporal and temporal-spatial paths) can model complementary interaction relations.

In the last step, we use the actor interaction relation features to realize action recognition and group activity recognition. Up to now, the dimension of the relation feature matrix $ \hat X $ is $ T \times N \times D $. The first dimension of $ \hat X $ is pooled to obtain actor-level feature representation for individual action recognition. The second and third dimensions of $ \hat X $ are pooled to obtain group-level feature representation for GAR. We use two full connection layers in each path to building individual action recognition classifiers and group recognition classifiers. Finally, we fuse the classifier results of two paths by late fusion as the final decisions.

\subsection{Actor Interaction Relation learning}

\subsubsection{Spatial/Temporal Relation Modules}

To understand the actor's spatial-temporal evolution, we design the basic spatial-temporal interaction relation modeling modules by MLP. Unlike the previous work, we don’t use the relation matrix and self-attention mechanism but use the concise FC operation to implicitly model the interaction relation.

\textbf{Spatial Relation Learning Module:} To model actor spatial interaction relation in each frame, we propose the basic spatial relation module by MLP, called MLP-S. Specifically, we assume that $ {X^t} \in {R^{N \times D}} $ represents all the actor’s features in the \emph{t}-th frame. With $ {X^t} $ as input, the relation feature obtained by MLP-S is $ {\tilde X^t} \in {R^{N \times D}} $. The detailed steps of MLP-S are as follows: 

\begin{equation} \label{eq1}
{X^{'}} = LN({X^{t}})
\end{equation}

\begin{equation} \label{eq2}
{X^{''}} = Droptout(FC(Dropout(GeLU(FC({X^{'}})))))
\end{equation}

\begin{equation} \label{eq3}
{\tilde X^t} = {X^t} + {X^{''}}
\end{equation}

First, as shown in Eq. (\ref{eq1}), we use LayerNorm to normalize the feature $ {X^t} $ to obtain feature $ {X^{'}} $. Then, as shown in Eq. (\ref{eq2}), MLP block uses $ {X^{'}} $ as input to model the actor spatial interaction relation to obtain relation feature $ {X^{''}} $. The MLP module is only composed of FC, GeLU, and Dropout, which significantly simplifies the model’s complexity. Finally, as shown in Eq. (\ref{eq3}), to make our model easier to optimize, we fuse the original features $ {X^t} $ and features $ {X^{''}} $ by the residual connection. It is also worth noting that we do not use position encoding as the additional input. This is because that MLP-S is sensitive to the actor’s feature order in the feature matrix.

\textbf{Temporal Relation Learning Module:} To model the evolution of each actor's action, we also propose the basic temporal relation module by MLP, called MLP-T. Like the MLP-S, the process of MLP-T also follows the Eq. (\ref{eq1})-(\ref{eq3}). Differently, MLP-T takes the same actor's features in different frames as input. Specifically, we assume that $ {X^n} \in {R^{T \times D}} $ represents the \emph{n}-th actor’s features in all frames. In this way, MLP-T can capture the actor’s temporal interaction relation. Hence, taking $ {X^n} $ as input, the relation feature obtained after MLP-T modeling is $ {\tilde X^n} \in {R^{T \times D}} $. Similarly, we do not use temporal encoding as the additional input.

\subsubsection{Relation Feature Refining Module}

We can find that MLP-S and MLP-T only model the relation between different features in the same dimension without considering the relation between different dimensions in the same feature. Therefore, we propose a relation refining module to improve the feature’s expression ability.

Like MLP-S, the process of MLP-R also follows the Eq. (\ref{eq1})-(\ref{eq3}). Differently, MLP-R model the relation inside each feature. Specifically, we assume that $ {X^{t,n}}\in {R^D} $ represents the features of the \emph{n}-th actor in \emph{t}-th frame. In this way, MLP-R can model the relation in different dimensions to improve the expression ability of features. Therefore, taking $ {X^{t,n}} $ as input, the refining feature obtained after MLP-R modeling is $ {\tilde X^{t,n}} \in {R^D} $.

\subsubsection{Dual-path Actor Learning}

Our SRM and TRM are composed of the most basic modules. As shown in Eq. (\ref{eq4}), MLP-R is connected in series after the MLP-S to form the SRM. Similarly, as shown in Eq. (\ref{eq5}), MLP-R is connected in series after MLP-T to form TRM. Through the cooperation of basic modules, our method can obtain spatial-temporal features with stronger expression ability.

Then, as shown in Eq. (\ref{eq6}) and (\ref{eq7}), the dual-path of relation learning is constructed by SAMs and TRMs according to different spatial-temporal orders. ST (Spatial-Temporal) path is skilled at recognizing activities with the obvious spatial arrangement. TS (Temporal-Spatial) path is good at recognizing activities with obvious action evolution of each actor.

\begin{equation} \label{eq4}
SRM({X^t}) = MLP - R(MLP - S({X^t}))
\end{equation}

\begin{equation} \label{eq5}
TRM({X^n}) = MLP - R(MLP - T({X^n}))
\end{equation}

\begin{equation} \label{eq6}
ST(X) = TRMs((X + SRMs(X)))
\end{equation}

\begin{equation} \label{eq7}
TS(X) = SRMs((X + TRMs(X)))
\end{equation}

\subsection{Network Optimization}

To improve the training efficiency of the model, our model adopts a one-stage and end-to-end training strategy. The model includes two tasks, the main task is group activity recognition, and the subtask is individual action recognition. The goal of individual action recognition is to promote actor interaction relation learning. With the help of cross-entropy loss, the optimization process of our model is as follows:

\begin{equation} \label{eq8}
{L_{Class}} = {L_G}(\frac{{\bar y_{ST}^G + \bar y_{TS}^G + \bar y_{Scene}^G}}{3},{y^G}) + \lambda {L_I}(\frac{{\bar y_{ST}^I + \bar y_{TS}^I}}{2},{y^I})
\end{equation}

Where $ {L_{Class}} $, $ {L_G} $, and $ {L_I} $ denote the model’s classification loss and the cross-entropy classification loss of group activity recognition and individual action recognition, respectively. $ \bar y_{ST}^G $ and $ \bar y_{TS}^G $ denote the classification score of GAR in the ST path and TS path. In addition, we believe that scene information is beneficial to judge the group’s activity category. Therefore, to obtain scene features, we use SPP (Spatial Pyramid Pooling layer) to process the backbone’s output features. Then, we use the scene features to predict group activity categories. $ \bar y_{Scene}^G $ denotes the classification score of GAR by using the scene features. $ \bar y_{ST}^I $ and $ \bar y_{TS}^I $ denote the classification score of individual action recognition in the ST path and TS path. $ {y^G} $ and $ {y^I} $ denote the ground truth of group activity recognition and individual action recognition, respectively. $ \lambda $ is the hyper-parameter to balance the importance of two items.

\section{Experiment}

In this section, we validate our proposed method through experiments on two benchmarks. In section 4.1, we first introduce the details of benchmarks and evaluation metrics. In section 4.2, we introduce the implementation details of our model. In section 4.3, we verify the effectiveness of our proposed modules through a series of ablation studies and give the visualization results of some experiments. In section 4.4, we compare our method with the state-of-the-arts and deeply analyze the experiment results.

\subsection{Dataset and Evaluation Metric}

Dataset: Following \cite{10_wu2019learning,11_gavrilyuk2020actor,16_han2022dual,17_li2021groupformer,18_li2022learning}, we conduct experiments on two widely used benchmarks. (1) \textbf{Volleyball Dataset} consists of 4830 clips, including 3493 clips for training and 1337 clips for testing. The dataset is annotated with eight activity categories: right set, right spike, right pass, right win-point, left set, left spike, left pass, and left win-point. The actor’s bounding box and action category are annotated in the clip’s middle frame. The individual action categories include waiting, setting, digging, falling, spiking, blocking, jumping, moving, and standing. (2) \textbf{Collective Activity} includes 2481 clips of 44 videos. Following the same experimental setting of \cite{16_han2022dual,18_li2022learning}, we select 1/3 of the video clips for testing and the rest for training. The dataset is annotated with five activity categories: crossing, waiting, queuing, walking, and talking. Each clip’s bounding box and action category are annotated in the clip’s middle frame. The individual categories include NA, crossing, waiting, queuing, walking, and talking. The maximum number of individual labels determines the group activity label in each clip. Consistent with \cite{16_han2022dual,18_li2022learning}, we merge crossing and walking into moving.

Evaluation Metric: In both datasets, we following \cite{18_li2022learning} use MCA (Multi-Class Accuracy) and MPCA (Mean Per-Class Accuracy) as the evaluation metrics. The MCA denotes the percentage of correct in all categories. The MPCA denotes the average accuracy of all categories. Moreover, we also following \cite{18_li2022learning,48_yuan2021learning} use the confusion matrix to analyze the model’s performance in each category in both datasets.

\subsection{Implementation Details}

On the Volleyball dataset, we following \cite{18_li2022learning} use Inception-v3 as the backbone to extract features of each sampled frame. Like \cite{10_wu2019learning,18_li2022learning}, we randomly select three frames for training and select nine frames for testing. The shape of each frame is 720×1280. We use a 256-dimensional feature vector for each actor. In addition, we use the Adamw optimizer to optimizer our model with the warmup strategy. We train our model with 150 epochs, where the epoch of warmup is 30, and the base learning rate is 0.0002. We set $ \lambda  = 1 $.

On the Collective dataset, we follow similar details to the Volleyball dataset. Differently, the shape of each frame is 480×720. We train our model with 80 epochs, where the epoch of warmup is 30, and the base learning rate is 0.0001. Finally, all our experiments are conducted on the PyTorch deep learning framework on two Nvidia GeForce RTX 3090 GPUs.

\subsection{Ablation Studies}

In this section, we first verify the effectiveness of our proposed relation modules, including MLP-S, MLP-T, and MLP-R. Then, we analyze the influence of scene information and the number of relation modules on the model’s performance. Next, we analyze the complexity of the relation modules. Finally, we visually visualize the group representations in different paths.

\begin{table}[htbp]\normalsize 
\begin{center}
\renewcommand\arraystretch{1.7}
\renewcommand\tabcolsep{10pt}
\begin{flushleft}
\caption{The ablation study of our proposed modules on the volleyball dataset. Single-Path-ST and Single-Path-TS are composed of SRM and TRM in different orders. Dual-Path-ST-TS denotes that we fuse the classification scores of the Single-Path-ST and Single-Path-TS.}
\label{Table 1}
\end{flushleft}
\begin{tabular}{lcc}
\hline
\multicolumn{2}{c}{Path}                                                                  & \begin{tabular}[c]{@{}c@{}}Group Activity\\ (MCA)\end{tabular} \\ \hline
\multirow{2}{*}{Single-Path} & ST                                                         & 92.2                                                           \\
                             & TS                                                         & 92.3                                                           \\ \hline
\multirow{2}{*}{Dual-Path}   & \begin{tabular}[c]{@{}c@{}}ST-TS-w/o-\\ MLP-R\end{tabular} & 91.8                                                           \\
                             & \textbf{ST-TS}                                             & \textbf{93.3}                                                  \\ \hline
\end{tabular}
\end{center}
\end{table}

\textbf{1) Ablation Studies for relation modeling:} To verify the effectiveness of our method, we conduct a couple of ablation studies, as shown in Table \ref{Table 1}. Like \cite{16_han2022dual}, we first verify the impact of different spatial-temporal relation feature extraction orders on the model’s performance. Single-Path-ST and Single-Path-TS are composed of two consecutive modules by using SRM and TRM in different orders. We can find that both methods obtain effective performance (92.2 and 92.3), respectively. This indicates that both Single-Path-ST and Single-Path-TS can model the spatial-temporal relation, which makes the relation features contain more abundant information. Secondly, we fuse the classification scores of Single-Path-ST and Single-Path-TS and denote them as Dual-Path-ST-TS. Dual-Path-ST-TS outperforms Single-Path-ST and Single-Path-TS by 1.1 (93.3 vs.92.2) and 1.0 (93.3 vs. 92.3), respectively. This shows that Single-Path-ST and Single-Path-TS can learn the complementary relation features, and the model’s performance can be improved by fusing these relation features. Thirdly, we also verify the effectiveness of MLP-R on relation refinement. We remove the MLP-R module in the Dual-ST-TS and denote it as Dual-ST-TS-w/o-MLP-R. Compared with the Dual-ST-TS-w/o-MLP-R, Dual-ST-TS is increased by 1.5 (93.3 vs. 91.8), which shows MLP-R can refine the relation features to improve the ability of relation expression.

\begin{table}[htbp]\normalsize 
\begin{center}
\renewcommand\arraystretch{1.7}
\renewcommand\tabcolsep{10pt}
\begin{flushleft}
\caption{The impact of scene information on the model’s performance.}\label{Table 2}
\end{flushleft}
\begin{tabular}{cc}
\hline
\begin{tabular}[c]{@{}c@{}}Scene\\ Information\end{tabular} & \begin{tabular}[c]{@{}c@{}}Group Activity\\ (MCA)\end{tabular} \\ \hline
ST-TS-w/o-SI                                                & 93.3                                                           \\
\textbf{ST-TS}                                                       & \textbf{93.7}                                                           \\ \hline
\end{tabular}
\end{center}
\end{table}

\textbf{2) Ablation Studies for Scene Information:} We verify the impact of scene information on the model’s performance, as shown in Table \ref{Table 2}. We remove the scene information in the Dual-ST-TS and denote it as Dual-ST-TS-w/o-SI. Compared with Dual-ST-TS-w/o-SI, Dual-ST-TS is increased by 0.4 (93.7 vs. 93.3). This proves scene information can further improve the model’s performance, which shows that group activities are closely related to the scene.

\begin{table}[htbp]\normalsize 
\begin{center}
\renewcommand\arraystretch{1.7}
\renewcommand\tabcolsep{10pt}
\begin{flushleft}
\caption{The impact of relation module’s number on the model’s performance.}\label{Table 3}
\end{flushleft}
\begin{tabular}{ccc}
\hline
Dual-Path      & Block Number & \begin{tabular}[c]{@{}c@{}}Group Activity\\ (MCA)\end{tabular} \\ \hline
\textbf{ST-TS} & \textbf{1}   & \textbf{93.7}                                                  \\
ST-TS          & 2            & 93.3                                                           \\
ST-TS          & 3            & 93.6                                                           \\
ST-TS          & 4            & 93.3                                                           \\
ST-TS          & 5            & 93.4                                                           \\ \hline
\end{tabular}
\end{center}
\end{table}

\textbf{3) Ablation Studies for Different Numbers of Relation Modules:} We verify the impact of multi-layer interaction relations on the model’s performance, as shown in Table \ref{Table 3}. We stack multiple SRMs and multiple TRMs in succession to form the multi-layer SRM and TRM, respectively. In Table \ref{Table 3}, the Block Number is the number of SRM and TRM in each multi-layer SRM and TRM, respectively. We denote each experiment as Dual-Path-ST-TS-X (X is the Block Number). We can find that Dual-Path-ST-TS-1 achieves the best result. Compared with the Dual-Path-ST-TS-3, Dual-Path-ST-TS-1 is increased by 0.1 (93.7 vs. 93.6). This proves that our proposed modules have a stronger ability for relation learning. Only need one layer can model rich relation features. Meanwhile, fewer layers make the model easier to optimize.

\begin{table}[htbp]\normalsize 
\begin{center}
\renewcommand\arraystretch{1.7}
\renewcommand\tabcolsep{7pt}
\begin{flushleft}
\caption{The ablation studies of complexity on GCN, Transformer, and MLP-AIR.}\label{Table 4}
\end{flushleft}
\begin{tabular}{cccc}
\hline
Method                       & Param          & Flops            & \begin{tabular}[c]{@{}c@{}}Group Activity\\ (MCA)\end{tabular} \\ \hline
\multirow{5}{*}{GCN}         & 0.79M          & 84.96M           & 93.2                                                           \\
                             & 1.58M          & 169.84M          & 93.3                                                           \\
                             & \textbf{2.37M} & \textbf{254.80M} & \textbf{93.4}                                                  \\
                             & 3.15M          & 339.76M          & 93.0                                                           \\
                             & 4.72M          & 509.60M          & 93.3                                                           \\ \hline
\multirow{5}{*}{Transformer} & 1.58M          & 175.44M          & 93.2                                                           \\
                             & \textbf{3.16M} & \textbf{350.88M} & \textbf{93.6}                                                  \\
                             & 4.72M          & 526.32M          & 93.4                                                           \\
                             & 6.32M          & 701.76M          & 93.3                                                           \\
                             & 7.88M          & 877.20M          & 92.7                                                           \\ \hline
\multirow{5}{*}{MLP-AIR}     & \textbf{0.53M} & \textbf{58.52M}  & \textbf{93.7}                                                  \\
                             & 1.05M          & 117.04M          & 93.3                                                           \\
                             & 1.58M          & 175.52M          & 93.6                                                           \\
                             & 2.11M          & 234.00M          & 93.3                                                           \\
                             & 2.64M          & 292.52M          & 93.4                                                           \\ \hline
\end{tabular}
\end{center}
\end{table}

\textbf{4) Ablation Studies for the Relation Module’s Complexity:} To verify the complexity of our method, called MLP-AIR, we compare it with GCN and Transformer under a unified framework. We implement the GCN and Transformer based on the \cite{10_wu2019learning} and \cite{16_han2022dual}. We conduct experiments under different model complexities for each method and count the parameters and FLOPs of relation modules, as shown in Table \ref{Table 4}. Compared with the best performance of GCN, MLP-AIR only uses about 22$\%$ parameters (0.53M vs. 2.37M) and 23$\%$ FLOPs (58.52M vs. 254.80M) to get higher performance (93.7 vs. 93.4). Compared with the best performance of Transformer, MLP-AIR is better than Transformer (93.7 vs. 93.6). It is worth noting that MLP-AIR only uses 17$\%$ parameters (0.53M vs. 3.16M) and FLOPs (58.52M vs. 350.88M) of the Transformer. Through the above analysis, we think that MLP-AIR has two advantages. Firstly, MLP-AIR can effectively model actor interaction relation and gain competitive results. Secondly, MLP-AIR is a simple relation modeling method with low complexity.

\begin{figure*}[Figure 4]
    \centering
    \subfigure{
        \includegraphics[width=1\textwidth]{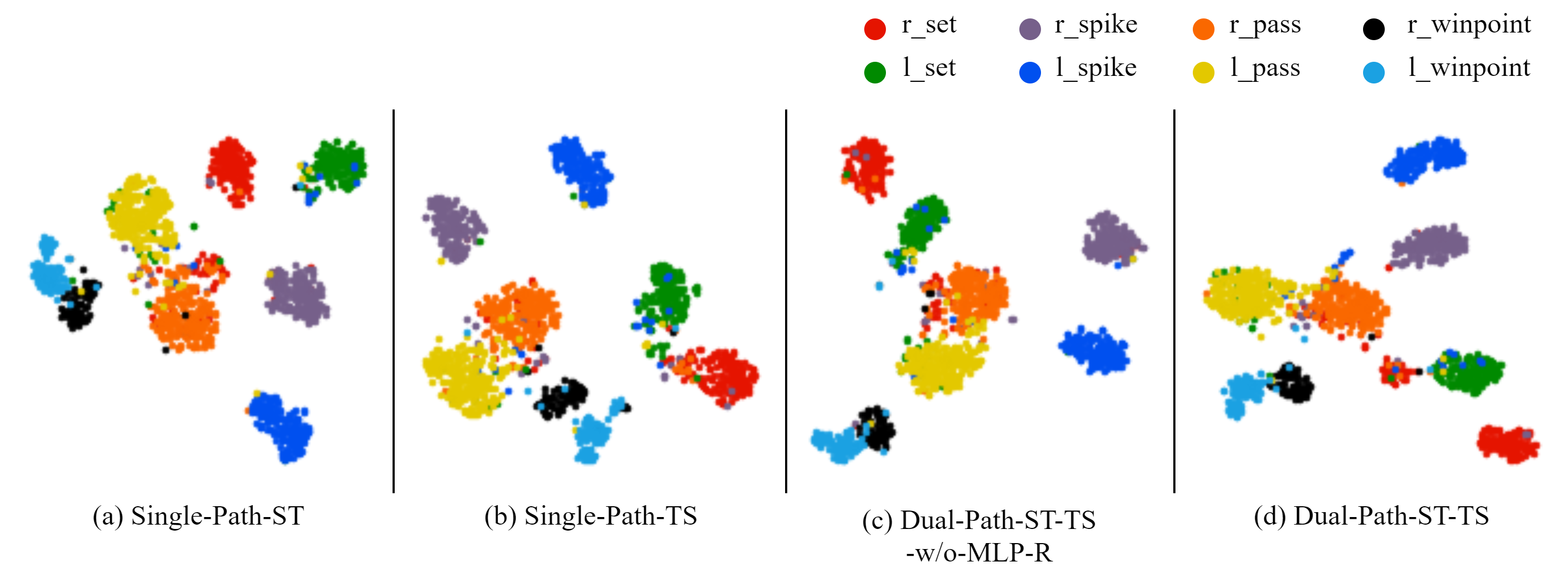}
    }
\begin{flushleft}
\caption{The visualization of group representations extracted from different paths of our method. "l$\_$" and "r$\_$" are abbreviations of "left" and "right" in group activity labels for the Volleyball dataset.}\label{Figure 4}
\end{flushleft}
\end{figure*}

\begin{table*}[h]\normalsize 
\begin{center}
\renewcommand\arraystretch{1.7}
\renewcommand\tabcolsep{15pt}
\begin{flushleft}
\caption{Comparison with state-of-the-art methods on Volleyball dataset. "$\_$" denotes that the original paper does not provide the result. All methods only use RGB images as input. Former is short for Transformer.}\label{Table 5}
\end{flushleft}
\begin{tabular}{cccccc}
\hline
Method                  & Backbone     & \begin{tabular}[c]{@{}c@{}}Relation\\ Learning\\ Method\end{tabular} & \begin{tabular}[c]{@{}c@{}}Individual\\ Action\\ (MCA)\end{tabular} & \begin{tabular}[c]{@{}c@{}}Group\\ Activity\\ (MCA)\end{tabular} & \begin{tabular}[c]{@{}c@{}}Group\\ Activity\\ (MPCA)\end{tabular} \\ \hline
SSU\cite{45_bagautdinov2017social}(2017)       & Inception-v3 & RNN                                                                  & 81.8                                                                & -                                                                & 90.6                                                              \\
PC-TDM\cite{28_yan2018participation}(2018)    & AlexNet      & RNN                                                                  & -                                                                   & 87.7                                                             & 88.1                                                              \\
ARG\cite{10_wu2019learning}(2019)       & Inception-v3 & Graph                                                                & 83.0                                                                & 92.5                                                             & -                                                                 \\
StageNet\cite{27_qi2019stagnet}(2020)  & VGG-16       & Graph                                                                & -                                                                   & 89.3                                                             & -                                                                 \\
HiGCN\cite{13_9241410}(2020)     & ResNet-18    & Graph                                                                & -                                                                   & 91.4                                                             & 92.0                                                              \\
PRL\cite{50_hu2020progressive}(2020)       & VGG-16       & Graph                                                                & -                                                                   & 91.4                                                             & 91.8                                                              \\
AFormer\cite{11_gavrilyuk2020actor}(2020)   & I3D          & Former                                                               & -                                                                   & 91.4                                                             & -                                                                 \\
DIN\cite{15_yuan2021spatio}(2021)       & VGG-16       & Graph                                                                & -                                                                   & 93.6                                                             & 93.8                                                              \\
P$^2$CTDM\cite{51_yan2021position}(2021)    & Inception-v3 & RNN                                                                  & -                                                                   & 91.8                                                             & 92.7                                                              \\
TCE+STBiP\cite{48_yuan2021learning}(2021) & Inception-v3 & Graph                                                                & -                                                                   & 93.3                                                             & 93.4                                                              \\
ASTFormer\cite{18_li2022learning}(2022) & Inception-v3 & Former                                                               & -                                                                   & 93.5                                                             & 93.9                                                              \\
MLP-AIR                 & Inception-v3 & MLP                                                                  & 82.9                                                                & \textbf{93.7}                                                             & \textbf{93.9}                                                              \\ \hline
\end{tabular}
\end{center}
\end{table*}

\begin{table*}[htbp]\normalsize 
\begin{center}
\renewcommand\arraystretch{1.7}
\renewcommand\tabcolsep{10pt}
\begin{flushleft}
\caption{Comparison with state-of-the-art methods on Collective dataset. All methods only use RGB images as input.}\label{Table 6}
\end{flushleft}
\begin{tabular}{ccccc}
\hline
Method                  & Backbone          & \begin{tabular}[c]{@{}c@{}}Relation\\ Learning\\ Method\end{tabular} & \begin{tabular}[c]{@{}c@{}}Group\\ Activity\\ (MCA)\end{tabular} & \begin{tabular}[c]{@{}c@{}}Group\\ Activity\\ (MPCA)\end{tabular} \\ \hline
Recurrent\cite{52_wang2017recurrent}(2017) & AlexNet,GoogleNet & RNN                                                                  & -                                                                & 89.4                                                              \\
PC-TDM\cite{28_yan2018participation}(2018)    & AlexNet           & RNN                                                                  & -                                                                & 92.2                                                              \\
ARG\cite{10_wu2019learning}(2019)       & Inception-v3      & Graph                                                                & 91.0                                                             & -                                                                 \\
StageNet\cite{27_qi2019stagnet}(2020)  & VGG-16            & Graph                                                                & 89.1                                                             & -                                                                 \\
HiGCN\cite{13_9241410}(2020)     & ResNet-18         & Graph                                                                & 93.4                                                             & 93.0                                                              \\
PRL\cite{50_hu2020progressive}(2020)       & VGG-16            & Graph                                                                & -                                                                & 93.8                                                              \\
DIN\cite{15_yuan2021spatio}(2021)       & VGG-16            & Graph                                                                & -                                                                & 95.9                                                              \\
P$^2$CTDM\cite{51_yan2021position}(2021)    & Inception-v3      & RNN                                                                  & -                                                                & 94.1                                                              \\
TCE+STBiP\cite{48_yuan2021learning}(2021) & Inception-v3      & Graph                                                                & -                                                                & 95.1                                                              \\
ASTFormer\cite{18_li2022learning}(2022) & Inception-v3      & Former                                                               & 95.7                                                             & 95.3                                                              \\
MLP-AIR                 & Inception-v3      & MLP                                                                  & \textbf{97.1}                                                             & \textbf{95.8}                                                              \\ \hline
\end{tabular}
\end{center}
\end{table*}

\textbf{5) The Visualization of Group Representations:} To visually display the group representation extracted from different paths in Table \ref{Table 1}, we use t-SNE \cite{49_van2008visualizing} to map the group representation to two-dimensional space for visualization, as shown in Fig. \ref{Figure 4}. We find that group representations of different categories have a significant clustering effect, as shown in each sub-figure of Fig. \ref{Figure 4}. This indicates that our proposed MLP-based method can model specific actor interaction relations of each category. Compared with Single-Path-ST and Single-Path-TS, Dual-Path-ST-TS has a more significant clustering effect in categories "r$\_$pass" and "r$\_$winpoint". This proves that Dual-Path-ST-TS can fuse the complementary information of both paths to improve the model’s performance. Moreover, compared with Dual-Path-ST-TS-w/o-MLP-R, Dual-Path-ST-TS significantly improve the classification performance of "r$\_$pass" and "l$\_$pass". It effectively reduces the confusion between "r$\_$pass" and "r$\_$set", "l$\_$pass" and "r$\_$pass", which shows that MLP-R can further refine relations features to improve the expression ability of features.

\begin{figure*}[Figure 5]
    \centering
    \subfigure{
        \includegraphics[width=1\textwidth]{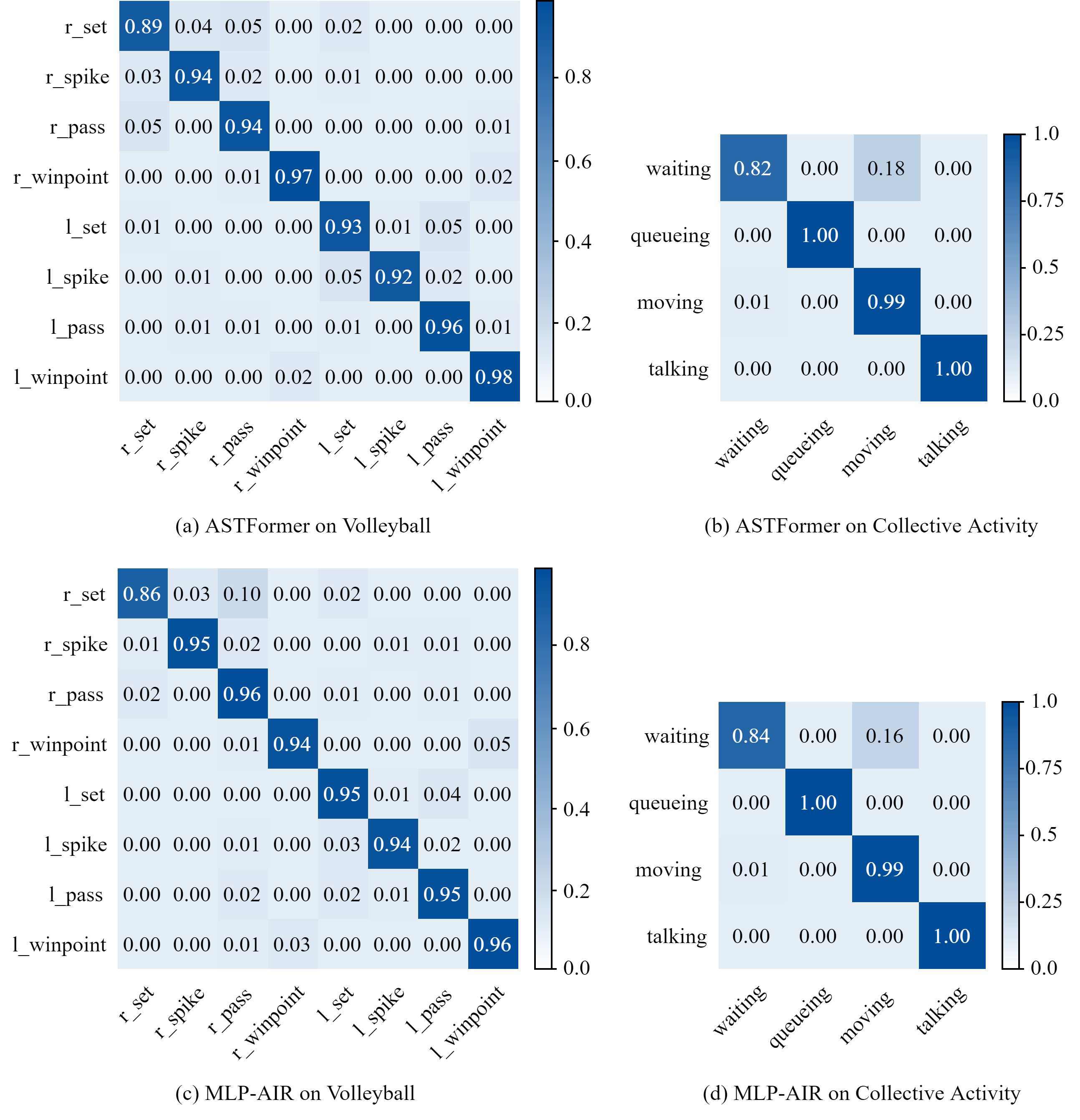}
    }
\begin{flushleft}
\caption{The confusion matrixes of ASTFormer and MLP-AIR on Volleyball and Collective Activity datasets. "l$\_$" and "r$\_$" are abbreviations of "left" and "right" in group activity labels for the Volleyball dataset.}\label{Figure 5}
\end{flushleft}
\end{figure*}

\subsection{Comparison with the State-of-the-Arts}

\textbf{1) Quantitative Analysis on Volleyball Dataset:} We compare different types of relation learning methods widely, including Transformer-based methods \cite{11_gavrilyuk2020actor,18_li2022learning}, Graph-based methods \cite{10_wu2019learning,13_9241410,15_yuan2021spatio,27_qi2019stagnet,48_yuan2021learning,50_hu2020progressive}, and RNN-based methods \cite{28_yan2018participation,45_bagautdinov2017social,51_yan2021position}. The details of the comparison are shown in Table \ref{Table 5}. We find that MLP-AIR achieves better performance. ASTFormer, TCE+STBiP, and P$^2$CTDM use the Transformer, Graph, and RNN as the relation modeling methods, respectively. Compared with ASTFormer, MLP-AIR is increased by 0.2 (93.7 vs. 93.5) under MAC and is equivalent (93.9 vs. 93.9) under MPAC. This shows that MLP-AIR without the self-attention mechanism has excellent spatial-temporal relation learning ability like Transformer-based methods. Moreover, MLP-AIR has a lower complexity for interaction relation modeling. In attention, MLP-AIR also outperforms TCE+STBiP and P$^2$CTDM by a significant margin. This shows that MLP-AIR also has a competitive relation modeling ability compared to other relation learning methods.

As shown in Fig. \ref{Figure 5}, through the confusion matrix, we conduct a more detailed analysis of MLP-AIR and ASTFormer. We first further analyze the performance of MLP-AIR in different categories. We find that the classification accuracies in all categories except "r$\_$set" exceed 94$\%$. This indicates that the spatial-temporal interaction relations modeled by MLP-AIR on different categories have significant differences, which can effectively improve the model’s classification performance. In addition, the accuracy of "r$\_$set" is lower than that of other categories. The main confusion is between "r$\_$set" and "r$\_$pass". We find that some instances of both categories are quite similar, making it difficult to distinguish them \cite{18_li2022learning,48_yuan2021learning}. Secondly, we also analyze the confusion matrix of ASTFormer. We find that the classification performance of ASTFormer outperformers MLP-AIR in half of all categories. We think that ASTFormer refines the person-level features by the action-centric aggregation to remove the redundant information, which is conducive to classifying similar group activities, such as "r$\_$set" and "r$\_$pass", "r$\_$winpoint" and "l$\_$winpoint".

\textbf{2) Quantitative Analysis on Collective Activity Dataset:} Table \ref{Table 6} shows the comparison results between MLP-AIR and other methods. It can be seen that MLP-AIR still achieves competitive performance. Compared with ASTFormer, MLP-AIR is improved by 1.4 (97.1 vs. 95.7) and 0.5 (95.8 vs. 95.3) under both metrics, respectively. MLP-AIR also outperforms TCE+STBiP and P$^2$CTDM. These results demonstrate the effectiveness of MLP-AIR for actor relation modeling again. We further find that the accuracy of DIN, PRL, and StageNet does not outperform MLP-AIR even though they use a more powerful backbone. This shows that interaction relation modeling is crucial for GAR, and MLP-AIR is a simple and efficient method for interaction relation modeling.

Through the confusion matrix, we also find that MLP-AIR obtains impressive performance in "queueing", "moving", and "talking" due to its significant ability of spatial-temporal relation modeling, as shown in Fig. \ref{Figure 5}. However, "waiting" is easily misclassified as "moving". To overcome this problem, we should further try to explore the scene evolution information around the object. Moreover, MLP-AIR outperforms ASTFormer in all categories in this dataset, which shows the effectiveness of relation modeling by MLP-AIR.

\section{Conclusion}

In this paper, we propose a novel MLP-based method to model actor spatial-temporal interaction relation and further refine the relation feature to improve the expression ability of features. To our knowledge, we are the first to propose an MLP-based method for GAR. Through a couple of ablation studies and visualization results, we prove the effectiveness of the proposed modules. Compared with the state-of-art on two benchmarks, our method achieves competitive performance. Most importantly, compared with the most commonly used relation modeling methods, such as GCNs and Transformers, our method can effectively model the actor interaction relation with a low model complexity.

\section{Acknowledgments}

This work was supported partly by the National Natural Science Foundation of China (Grant No. 62173045, 61673192), partly by the Fundamental Research Funds for the Central Universities (Grant No. 2020XD-A04-3), and the Natural Science Foundation of Hainan Province (Grant No. 622RC675).

%% Loading bibliography style file
% \bibliographystyle{cas-model2-names}
\bibliographystyle{model1-num-names}
% \bibliographystyle{unsrt}
% Loading bibliography database
\bibliography{cas-refs}

\begin{thebibliography}{52}
\expandafter\ifx\csname natexlab\endcsname\relax\def\natexlab#1{#1}\fi
\providecommand{\url}[1]{\texttt{#1}}
\providecommand{\href}[2]{#2}
\providecommand{\path}[1]{#1}
\providecommand{\DOIprefix}{doi:}
\providecommand{\ArXivprefix}{arXiv:}
\providecommand{\URLprefix}{URL: }
\providecommand{\Pubmedprefix}{pmid:}
\providecommand{\doi}[1]{\href{http://dx.doi.org/#1}{\path{#1}}}
\providecommand{\Pubmed}[1]{\href{pmid:#1}{\path{#1}}}
\providecommand{\bibinfo}[2]{#2}
\ifx\xfnm\relax \def\xfnm[#1]{\unskip,\space#1}\fi
%Type = Article
\bibitem[{Deng et~al.(2022)Deng, Wang, Li, and Lin}]{1_Deng2022Summarization}
\bibinfo{author}{H.~Deng}, \bibinfo{author}{C.~Wang}, \bibinfo{author}{C.~Li},
  \bibinfo{author}{X.~Lin},
\newblock \bibinfo{title}{Summarization of group activity recognition
  algorithms based on deep learning frame},
\newblock \bibinfo{journal}{Acta Electronica Sinica} \bibinfo{volume}{50}
  (\bibinfo{year}{2022}) \bibinfo{pages}{2018--2036}.
%Type = Inproceedings
\bibitem[{Mahsa et~al.(2020)Mahsa, Alireza, Fatemeh, Javen, Ian, and
  Hamid}]{2_ehsanpour2020joint}
\bibinfo{author}{E.~Mahsa}, \bibinfo{author}{A.~Alireza},
  \bibinfo{author}{S.~Fatemeh}, \bibinfo{author}{S.~Javen},
  \bibinfo{author}{R.~Ian}, \bibinfo{author}{R.~Hamid},
\newblock \bibinfo{title}{Joint learning of social groups, individuals action
  and sub-group activities in videos},
\newblock in: \bibinfo{booktitle}{Proceedings of the European Conference on
  Computer Vision}, \bibinfo{year}{2020}, pp. \bibinfo{pages}{177--195}.
%Type = Article
\bibitem[{Liu et~al.(2022)Liu, Zhao, Kin-Man, and Kong}]{3_liu2022visual}
\bibinfo{author}{T.~Liu}, \bibinfo{author}{R.~Zhao},
  \bibinfo{author}{L.~Kin-Man}, \bibinfo{author}{J.~Kong},
\newblock \bibinfo{title}{Visual-semantic graph neural network with
  pose-position attentive learning for group activity recognition},
\newblock \bibinfo{journal}{Neurocomputing} \bibinfo{volume}{491}
  (\bibinfo{year}{2022}) \bibinfo{pages}{217--231}.
%Type = Article
\bibitem[{Lu et~al.(2018)Lu, Di, Lu, Zhang, and Wang}]{4_LU2018195}
\bibinfo{author}{L.~Lu}, \bibinfo{author}{H.~Di}, \bibinfo{author}{Y.~Lu},
  \bibinfo{author}{L.~Zhang}, \bibinfo{author}{S.~Wang},
\newblock \bibinfo{title}{A two-level attention-based interaction model for
  multi-person activity recognition},
\newblock \bibinfo{journal}{Neurocomputing} \bibinfo{volume}{322}
  (\bibinfo{year}{2018}) \bibinfo{pages}{195--205}.
%Type = Article
\bibitem[{Wang et~al.(2021)Wang, Shao, Huang, Lu, Zhang, and
  Lv}]{5_WANG2021265}
\bibinfo{author}{J.~Wang}, \bibinfo{author}{Z.~Shao},
  \bibinfo{author}{X.~Huang}, \bibinfo{author}{T.~Lu},
  \bibinfo{author}{R.~Zhang}, \bibinfo{author}{X.~Lv},
\newblock \bibinfo{title}{Spatial–temporal pooling for action recognition in
  videos},
\newblock \bibinfo{journal}{Neurocomputing} \bibinfo{volume}{451}
  (\bibinfo{year}{2021}) \bibinfo{pages}{265--278}.
%Type = Article
\bibitem[{Zheng et~al.(2022)Zheng, Gong, Lu, and Li}]{6_ZHENG202210}
\bibinfo{author}{X.~Zheng}, \bibinfo{author}{T.~Gong}, \bibinfo{author}{X.~Lu},
  \bibinfo{author}{X.~Li},
\newblock \bibinfo{title}{Human action recognition by multiple spatial clues
  network},
\newblock \bibinfo{journal}{Neurocomputing} \bibinfo{volume}{483}
  (\bibinfo{year}{2022}) \bibinfo{pages}{10--21}.
%Type = Article
\bibitem[{Wu et~al.(2022)Wu, Song, Yue, Wang, Xiao, and Liu}]{7_WU2022358}
\bibinfo{author}{H.~Wu}, \bibinfo{author}{C.~Song}, \bibinfo{author}{S.~Yue},
  \bibinfo{author}{Z.~Wang}, \bibinfo{author}{J.~Xiao},
  \bibinfo{author}{Y.~Liu},
\newblock \bibinfo{title}{Dynamic video mix-up for cross-domain action
  recognition},
\newblock \bibinfo{journal}{Neurocomputing} \bibinfo{volume}{471}
  (\bibinfo{year}{2022}) \bibinfo{pages}{358--368}.
%Type = Article
\bibitem[{Yue et~al.(2022)Yue, Tian, and Du}]{8_YUE2022287}
\bibinfo{author}{R.~Yue}, \bibinfo{author}{Z.~Tian}, \bibinfo{author}{S.~Du},
\newblock \bibinfo{title}{Action recognition based on rgb and skeleton data
  sets: A survey},
\newblock \bibinfo{journal}{Neurocomputing} \bibinfo{volume}{512}
  (\bibinfo{year}{2022}) \bibinfo{pages}{287--306}.
%Type = Article
\bibitem[{Perez et~al.(2022)Perez, Liu, and Kot}]{9_PEREZ2022108360}
\bibinfo{author}{M.~Perez}, \bibinfo{author}{J.~Liu}, \bibinfo{author}{A.~C.
  Kot},
\newblock \bibinfo{title}{Skeleton-based relational reasoning for group
  activity analysis},
\newblock \bibinfo{journal}{Pattern Recognition} \bibinfo{volume}{122}
  (\bibinfo{year}{2022}) \bibinfo{pages}{108360}.
%Type = Inproceedings
\bibitem[{Wu et~al.(2019)Wu, Wang, Wang, Guo, and Wu}]{10_wu2019learning}
\bibinfo{author}{J.~Wu}, \bibinfo{author}{L.~Wang}, \bibinfo{author}{L.~Wang},
  \bibinfo{author}{J.~Guo}, \bibinfo{author}{G.~Wu},
\newblock \bibinfo{title}{Learning actor relation graphs for group activity
  recognition},
\newblock in: \bibinfo{booktitle}{Proceedings of the IEEE Conference on
  computer vision and pattern recognition}, \bibinfo{year}{2019}, pp.
  \bibinfo{pages}{9964--9974}.
%Type = Inproceedings
\bibitem[{Kirill et~al.(2020)Kirill, Ryan, Javan, and
  GM}]{11_gavrilyuk2020actor}
\bibinfo{author}{G.~Kirill}, \bibinfo{author}{S.~Ryan},
  \bibinfo{author}{M.~Javan}, \bibinfo{author}{S.~C. GM},
\newblock \bibinfo{title}{Actor-transformers for group activity recognition},
\newblock in: \bibinfo{booktitle}{Proceedings of the IEEE Conference on
  Computer Vision and Pattern Recognition}, \bibinfo{year}{2020}, pp.
  \bibinfo{pages}{839--848}.
%Type = Inproceedings
\bibitem[{Zhou et~al.(2022)Zhou, Asim, Aviv, Geng, Farley, Zhao, Liu, Mubbasir,
  and Peter}]{12_zhou2021composer}
\bibinfo{author}{H.~Zhou}, \bibinfo{author}{K.~Asim},
  \bibinfo{author}{S.~Aviv}, \bibinfo{author}{S.~Geng},
  \bibinfo{author}{L.~Farley}, \bibinfo{author}{L.~Zhao},
  \bibinfo{author}{T.~Liu}, \bibinfo{author}{K.~Mubbasir},
  \bibinfo{author}{G.~H. Peter},
\newblock \bibinfo{title}{Composer: Compositional learning of group activity in
  videos},
\newblock in: \bibinfo{booktitle}{Proceedings of the European Conference on
  Computer Vision}, \bibinfo{year}{2022}, pp. \bibinfo{pages}{1--47}.
%Type = Article
\bibitem[{Yan et~al.(2020{\natexlab{a}})Yan, Xie, Tang, Shu, and
  Tian}]{13_9241410}
\bibinfo{author}{R.~Yan}, \bibinfo{author}{L.~Xie}, \bibinfo{author}{J.~Tang},
  \bibinfo{author}{X.~Shu}, \bibinfo{author}{Q.~Tian},
\newblock \bibinfo{title}{Higcin: Hierarchical graph-based cross inference
  network for group activity recognition},
\newblock \bibinfo{journal}{IEEE Transactions on Pattern Analysis and Machine
  Intelligence}  (\bibinfo{year}{2020}{\natexlab{a}}) \bibinfo{pages}{1--14}.
%Type = Inproceedings
\bibitem[{Yan et~al.(2020{\natexlab{b}})Yan, Xie, Tang, Shu, and
  Tian}]{14_yan2020social}
\bibinfo{author}{R.~Yan}, \bibinfo{author}{L.~Xie}, \bibinfo{author}{J.~Tang},
  \bibinfo{author}{X.~Shu}, \bibinfo{author}{Q.~Tian},
\newblock \bibinfo{title}{Social adaptive module for weakly-supervised group
  activity recognition},
\newblock in: \bibinfo{booktitle}{Proceedings of the European Conference on
  Computer Vision}, \bibinfo{year}{2020}{\natexlab{b}}, pp.
  \bibinfo{pages}{208--224}.
%Type = Inproceedings
\bibitem[{Yuan et~al.(2021)Yuan, Ni, and Wang}]{15_yuan2021spatio}
\bibinfo{author}{H.~Yuan}, \bibinfo{author}{D.~Ni}, \bibinfo{author}{M.~Wang},
\newblock \bibinfo{title}{Spatio-temporal dynamic inference network for group
  activity recognition},
\newblock in: \bibinfo{booktitle}{Proceedings of the IEEE/CVF International
  Conference on Computer Vision}, \bibinfo{year}{2021}, pp.
  \bibinfo{pages}{7476--7485}.
%Type = Inproceedings
\bibitem[{Han et~al.(2022)Han, Zhang, Wang, Yan, Yao, Chang, and
  Yu}]{16_han2022dual}
\bibinfo{author}{M.~Han}, \bibinfo{author}{D.~J. Zhang},
  \bibinfo{author}{Y.~Wang}, \bibinfo{author}{R.~Yan},
  \bibinfo{author}{L.~Yao}, \bibinfo{author}{X.~Chang},
  \bibinfo{author}{Q.~Yu},
\newblock \bibinfo{title}{Dual-ai: dual-path actor interaction learning for
  group activity recognition},
\newblock in: \bibinfo{booktitle}{Proceedings of the IEEE/CVF conference on
  computer vision and pattern recognition}, \bibinfo{year}{2022}, pp.
  \bibinfo{pages}{2990--2999}.
%Type = Inproceedings
\bibitem[{Li et~al.(2021)Li, Cao, Liu, Yang, Liu, Hou, and
  Yi}]{17_li2021groupformer}
\bibinfo{author}{S.~Li}, \bibinfo{author}{Q.~Cao}, \bibinfo{author}{L.~Liu},
  \bibinfo{author}{K.~Yang}, \bibinfo{author}{S.~Liu},
  \bibinfo{author}{J.~Hou}, \bibinfo{author}{S.~Yi},
\newblock \bibinfo{title}{Groupformer: Group activity recognition with
  clustered spatial-temporal transformer},
\newblock in: \bibinfo{booktitle}{Proceedings of the IEEE/CVF International
  Conference on Computer Vision}, \bibinfo{year}{2021}, pp.
  \bibinfo{pages}{13668--13677}.
%Type = Inproceedings
\bibitem[{Li et~al.(2022)Li, Yang, Wu, Du, and Qiao}]{18_li2022learning}
\bibinfo{author}{W.~Li}, \bibinfo{author}{T.~Yang}, \bibinfo{author}{X.~Wu},
  \bibinfo{author}{X.~Du}, \bibinfo{author}{J.~Qiao},
\newblock \bibinfo{title}{Learning action-guided spatio-temporal transformer
  for group activity recognition},
\newblock in: \bibinfo{booktitle}{Proceedings of the 30th ACM International
  Conference on Multimedia}, \bibinfo{year}{2022}, pp.
  \bibinfo{pages}{2051--2060}.
%Type = Article
\bibitem[{Zhao et~al.(2021)Zhao, Wang, Tang, Luo, Zeng, and
  Zha}]{19_zhao2021battle}
\bibinfo{author}{Y.~Zhao}, \bibinfo{author}{G.~Wang},
  \bibinfo{author}{C.~Tang}, \bibinfo{author}{C.~Luo},
  \bibinfo{author}{W.~Zeng}, \bibinfo{author}{Z.~Zha},
\newblock \bibinfo{title}{A battle of network structures: An empirical study of
  cnn, transformer, and mlp},
\newblock \bibinfo{journal}{arXiv preprint arXiv:2108.13002}
  (\bibinfo{year}{2021}).
%Type = Article
\bibitem[{O et~al.(2021)O, Neil, Alexander, Lucas, Zhai, Thomas, Jessica,
  Andreas, Daniel, Jakob, Mario, and Alexey}]{20_tolstikhin2021mlp}
\bibinfo{author}{T.~I. O}, \bibinfo{author}{H.~Neil},
  \bibinfo{author}{K.~Alexander}, \bibinfo{author}{B.~Lucas},
  \bibinfo{author}{X.~Zhai}, \bibinfo{author}{U.~Thomas},
  \bibinfo{author}{Y.~Jessica}, \bibinfo{author}{S.~Andreas},
  \bibinfo{author}{K.~Daniel}, \bibinfo{author}{U.~Jakob},
  \bibinfo{author}{L.~Mario}, \bibinfo{author}{D.~Alexey},
\newblock \bibinfo{title}{Mlp-mixer: An all-mlp architecture for vision},
\newblock \bibinfo{journal}{Advances in neural information processing systems}
  \bibinfo{volume}{34} (\bibinfo{year}{2021}) \bibinfo{pages}{24261--24272}.
%Type = Inproceedings
\bibitem[{Zhang et~al.(2022)Zhang, Li, Wang, Chen, Shashwat, Yu, Liu, and
  Zheng}]{21_zhang2022morphmlp}
\bibinfo{author}{D.~J. Zhang}, \bibinfo{author}{K.~Li},
  \bibinfo{author}{Y.~Wang}, \bibinfo{author}{Y.~Chen},
  \bibinfo{author}{C.~Shashwat}, \bibinfo{author}{Q.~Yu},
  \bibinfo{author}{L.~Liu}, \bibinfo{author}{S.~M. Zheng},
\newblock \bibinfo{title}{Morphmlp: An efficient mlp-like backbone for
  spatial-temporal representation learning},
\newblock in: \bibinfo{booktitle}{Proceedings of the European Conference on
  Computer Vision}, \bibinfo{year}{2022}, pp. \bibinfo{pages}{230--248}.
%Type = Inproceedings
\bibitem[{Chen et~al.(2022)Chen, Xie, Ge, Chen, Ding, and
  Luo}]{22_chen2021cyclemlp}
\bibinfo{author}{S.~Chen}, \bibinfo{author}{E.~Xie}, \bibinfo{author}{C.~Ge},
  \bibinfo{author}{R.~Chen}, \bibinfo{author}{L.~Ding},
  \bibinfo{author}{P.~Luo},
\newblock \bibinfo{title}{Cyclemlp: A mlp-like architecture for dense
  prediction},
\newblock in: \bibinfo{booktitle}{International Conference on Learning
  Representations}, \bibinfo{year}{2022}, pp. \bibinfo{pages}{1--21}.
%Type = Inproceedings
\bibitem[{Felipe et~al.(2020)Felipe, Jorge, Marcelo, and
  Andres}]{23_borja2020deep}
\bibinfo{author}{B.-B.~L. Felipe}, \bibinfo{author}{A.-L. Jorge},
  \bibinfo{author}{S.-C. Marcelo}, \bibinfo{author}{F.-G. Andres},
\newblock \bibinfo{title}{Deep learning architecture for group activity
  recognition using description of local motions},
\newblock in: \bibinfo{booktitle}{International Joint Conference on Neural
  Networks}, \bibinfo{year}{2020}, pp. \bibinfo{pages}{1--8}.
%Type = Article
\bibitem[{Mokhtarzadeh et~al.(2018)Mokhtarzadeh, Ghadimi, and
  Ahmad}]{24_azar2018multi}
\bibinfo{author}{A.~S. Mokhtarzadeh}, \bibinfo{author}{A.~M. Ghadimi},
  \bibinfo{author}{N.~Ahmad},
\newblock \bibinfo{title}{A multi-stream convolutional neural network framework
  for group activity recognition},
\newblock \bibinfo{journal}{arXiv preprint arXiv:1812.10328}
  (\bibinfo{year}{2018}).
%Type = Inproceedings
\bibitem[{Takamasa et~al.(2017)Takamasa, Yasuhiro, Masakazu, and
  Tatsuya}]{25_tsunoda2017football}
\bibinfo{author}{T.~Takamasa}, \bibinfo{author}{K.~Yasuhiro},
  \bibinfo{author}{M.~Masakazu}, \bibinfo{author}{H.~Tatsuya},
\newblock \bibinfo{title}{Football action recognition using hierarchical lstm},
\newblock in: \bibinfo{booktitle}{Proceedings of the IEEE conference on
  computer vision and pattern recognition workshops}, \bibinfo{year}{2017}, pp.
  \bibinfo{pages}{99--107}.
%Type = Inproceedings
\bibitem[{Vignesh et~al.(2016)Vignesh, Jonathan, Sami, Alexander, Kevin, and
  Li}]{26_ramanathan2016detecting}
\bibinfo{author}{R.~Vignesh}, \bibinfo{author}{H.~Jonathan},
  \bibinfo{author}{A.-E.-H. Sami}, \bibinfo{author}{G.~Alexander},
  \bibinfo{author}{M.~Kevin}, \bibinfo{author}{F.~Li},
\newblock \bibinfo{title}{Detecting events and key actors in multi-person
  videos},
\newblock in: \bibinfo{booktitle}{Proceedings of the IEEE conference on
  computer vision and pattern recognition}, \bibinfo{year}{2016}, pp.
  \bibinfo{pages}{3043--3053}.
%Type = Article
\bibitem[{Qi et~al.(2019)Qi, Wang, Qin, Li, Luo, and Luc}]{27_qi2019stagnet}
\bibinfo{author}{M.~Qi}, \bibinfo{author}{Y.~Wang}, \bibinfo{author}{J.~Qin},
  \bibinfo{author}{A.~Li}, \bibinfo{author}{J.~Luo}, \bibinfo{author}{V.~G.
  Luc},
\newblock \bibinfo{title}{Stagnet: An attentive semantic rnn for group activity
  and individual action recognition},
\newblock \bibinfo{journal}{IEEE Transactions on Circuits and Systems for Video
  Technology} \bibinfo{volume}{30} (\bibinfo{year}{2019})
  \bibinfo{pages}{549--565}.
%Type = Inproceedings
\bibitem[{Yan et~al.(2018)Yan, Tang, Shu, Li, and
  Tian}]{28_yan2018participation}
\bibinfo{author}{R.~Yan}, \bibinfo{author}{J.~Tang}, \bibinfo{author}{X.~Shu},
  \bibinfo{author}{Z.~Li}, \bibinfo{author}{Q.~Tian},
\newblock \bibinfo{title}{Participation-contributed temporal dynamic model for
  group activity recognition},
\newblock in: \bibinfo{booktitle}{Proceedings of the 26th ACM international
  conference on Multimedia}, \bibinfo{year}{2018}, pp.
  \bibinfo{pages}{1292--1300}.
%Type = Inproceedings
\bibitem[{N and Max(2017)}]{29_kipf2016semi}
\bibinfo{author}{K.~T. N}, \bibinfo{author}{W.~Max},
\newblock \bibinfo{title}{Semi-supervised classification with graph
  convolutional networks},
\newblock in: \bibinfo{booktitle}{International Conference on Learning
  Representations}, \bibinfo{year}{2017}, pp. \bibinfo{pages}{1--14}.
%Type = Inproceedings
\bibitem[{Yan et~al.(2018)Yan, Xiong, and Lin}]{30_yan2018spatial}
\bibinfo{author}{S.~Yan}, \bibinfo{author}{Y.~Xiong}, \bibinfo{author}{D.~Lin},
\newblock \bibinfo{title}{Spatial temporal graph convolutional networks for
  skeleton-based action recognition},
\newblock in: \bibinfo{booktitle}{Proceedings of the AAAI conference on
  artificial intelligence}, \bibinfo{year}{2018}, pp.
  \bibinfo{pages}{7444--7452}.
%Type = Inproceedings
\bibitem[{Duan et~al.(2022)Duan, Zhao, Chen, Lin, and
  Dai}]{31_duan2022revisiting}
\bibinfo{author}{H.~Duan}, \bibinfo{author}{Y.~Zhao},
  \bibinfo{author}{K.~Chen}, \bibinfo{author}{D.~Lin},
  \bibinfo{author}{B.~Dai},
\newblock \bibinfo{title}{Revisiting skeleton-based action recognition},
\newblock in: \bibinfo{booktitle}{Proceedings of the IEEE Conference on
  Computer Vision and Pattern Recognition}, \bibinfo{year}{2022}, pp.
  \bibinfo{pages}{2969--2978}.
%Type = Article
\bibitem[{Sangjin et~al.(2022)Sangjin, Sangyeob, Juhyoung, and
  Hoijun}]{32_kim2022low}
\bibinfo{author}{K.~Sangjin}, \bibinfo{author}{K.~Sangyeob},
  \bibinfo{author}{L.~Juhyoung}, \bibinfo{author}{Y.~Hoijun},
\newblock \bibinfo{title}{A low-power graph convolutional network processor
  with sparse grouping for 3d point cloud semantic segmentation in mobile
  devices},
\newblock \bibinfo{journal}{IEEE Transactions on Circuits and Systems I:
  Regular Papers} \bibinfo{volume}{69} (\bibinfo{year}{2022})
  \bibinfo{pages}{1507--1518}.
%Type = Article
\bibitem[{Fei et~al.(2022)Fei, Yang, Chen, Li, Li, Ma, Hu, and
  Ma}]{33_fei2022comprehensive}
\bibinfo{author}{B.~Fei}, \bibinfo{author}{W.~Yang}, \bibinfo{author}{W.~Chen},
  \bibinfo{author}{Z.~Li}, \bibinfo{author}{Y.~Li}, \bibinfo{author}{T.~Ma},
  \bibinfo{author}{X.~Hu}, \bibinfo{author}{L.~Ma},
\newblock \bibinfo{title}{Comprehensive review of deep learning-based 3d point
  cloud completion processing and analysis},
\newblock \bibinfo{journal}{IEEE Transactions on Intelligent Transportation
  Systems} \bibinfo{volume}{23} (\bibinfo{year}{2022})
  \bibinfo{pages}{22862--22883}.
%Type = Inproceedings
\bibitem[{Ashish et~al.(2017)Ashish, Noam, Niki, Jakob, Llion, N, ukasz, and
  Illia}]{34_NIPS2017_3f5ee243}
\bibinfo{author}{V.~Ashish}, \bibinfo{author}{S.~Noam},
  \bibinfo{author}{P.~Niki}, \bibinfo{author}{U.~Jakob},
  \bibinfo{author}{J.~Llion}, \bibinfo{author}{G.~A. N},
  \bibinfo{author}{K.~ukasz}, \bibinfo{author}{P.~Illia},
\newblock \bibinfo{title}{Attention is all you need},
\newblock in: \bibinfo{booktitle}{Advances in Neural Information Processing
  Systems}, \bibinfo{year}{2017}, pp. \bibinfo{pages}{1--11}.
%Type = Inproceedings
\bibitem[{Alexey et~al.(2021)Alexey, Lucas, Alexander, Dirk, Zhai, Thomas,
  Mostafa, Matthias, Georg, Sylvain, Jakob, and Neil}]{35_dosovitskiy2020image}
\bibinfo{author}{D.~Alexey}, \bibinfo{author}{B.~Lucas},
  \bibinfo{author}{K.~Alexander}, \bibinfo{author}{W.~Dirk},
  \bibinfo{author}{X.~Zhai}, \bibinfo{author}{U.~Thomas},
  \bibinfo{author}{D.~Mostafa}, \bibinfo{author}{M.~Matthias},
  \bibinfo{author}{H.~Georg}, \bibinfo{author}{G.~Sylvain},
  \bibinfo{author}{U.~Jakob}, \bibinfo{author}{H.~Neil},
\newblock \bibinfo{title}{An image is worth 16x16 words: Transformers for image
  recognition at scale},
\newblock in: \bibinfo{booktitle}{International Conference on Learning
  Representations}, \bibinfo{year}{2021}, pp. \bibinfo{pages}{1--22}.
%Type = Inproceedings
\bibitem[{Wang et~al.(2021)Wang, Xie, Li, Fan, Song, Ding, Lu, Luo, and
  Shao}]{36_wang2021pyramid}
\bibinfo{author}{W.~Wang}, \bibinfo{author}{E.~Xie}, \bibinfo{author}{X.~Li},
  \bibinfo{author}{D.~Fan}, \bibinfo{author}{K.~Song},
  \bibinfo{author}{L.~Ding}, \bibinfo{author}{T.~Lu}, \bibinfo{author}{P.~Luo},
  \bibinfo{author}{L.~Shao},
\newblock \bibinfo{title}{Pyramid vision transformer: A versatile backbone for
  dense prediction without convolutions},
\newblock in: \bibinfo{booktitle}{Proceedings of the IEEE international
  conference on computer vision}, \bibinfo{year}{2021}, pp.
  \bibinfo{pages}{568--578}.
%Type = Inproceedings
\bibitem[{Aravind et~al.(2021)Aravind, Lin, Niki, Jonathon, Pieter, and
  Ashish}]{37_srinivas2021bottleneck}
\bibinfo{author}{S.~Aravind}, \bibinfo{author}{T.~Lin},
  \bibinfo{author}{P.~Niki}, \bibinfo{author}{S.~Jonathon},
  \bibinfo{author}{A.~Pieter}, \bibinfo{author}{V.~Ashish},
\newblock \bibinfo{title}{Bottleneck transformers for visual recognition},
\newblock in: \bibinfo{booktitle}{Proceedings of the IEEE conference on
  computer vision and pattern recognition}, \bibinfo{year}{2021}, pp.
  \bibinfo{pages}{16519--16529}.
%Type = Inproceedings
\bibitem[{Liu et~al.(2021)Liu, Lin, Cao, Hu, Wei, Zhang, Lin, and
  Guo}]{38_liu2021swin}
\bibinfo{author}{Z.~Liu}, \bibinfo{author}{Y.~Lin}, \bibinfo{author}{Y.~Cao},
  \bibinfo{author}{H.~Hu}, \bibinfo{author}{Y.~Wei},
  \bibinfo{author}{Z.~Zhang}, \bibinfo{author}{S.~Lin},
  \bibinfo{author}{B.~Guo},
\newblock \bibinfo{title}{Swin transformer: Hierarchical vision transformer
  using shifted windows},
\newblock in: \bibinfo{booktitle}{Proceedings of the IEEE international
  conference on computer vision}, \bibinfo{year}{2021}, pp.
  \bibinfo{pages}{10012--10022}.
%Type = Inproceedings
\bibitem[{Yang et~al.(2021)Yang, Li, Zhang, Dai, Xiao, Yuan, and
  Gao}]{39_yang2021focal}
\bibinfo{author}{J.~Yang}, \bibinfo{author}{C.~Li}, \bibinfo{author}{P.~Zhang},
  \bibinfo{author}{X.~Dai}, \bibinfo{author}{B.~Xiao},
  \bibinfo{author}{L.~Yuan}, \bibinfo{author}{J.~Gao},
\newblock \bibinfo{title}{Focal attention for long-range interactions in vision
  transformers},
\newblock in: \bibinfo{booktitle}{Advances in Neural Information Processing
  Systems}, \bibinfo{year}{2021}, pp. \bibinfo{pages}{30008--30022}.
%Type = Inproceedings
\bibitem[{Fan et~al.(2021)Fan, Xiong, Karttikeya, Li, Yan, Jitendra, and
  Christoph}]{40_fan2021multiscale}
\bibinfo{author}{H.~Fan}, \bibinfo{author}{B.~Xiong},
  \bibinfo{author}{M.~Karttikeya}, \bibinfo{author}{Y.~Li},
  \bibinfo{author}{Z.~Yan}, \bibinfo{author}{M.~Jitendra},
  \bibinfo{author}{F.~Christoph},
\newblock \bibinfo{title}{Multiscale vision transformers},
\newblock in: \bibinfo{booktitle}{Proceedings of the IEEE International
  Conference on Computer Vision}, \bibinfo{year}{2021}, pp.
  \bibinfo{pages}{6824--6835}.
%Type = Article
\bibitem[{Hou et~al.(2022)Hou, Jiang, Yuan, Cheng, Yan, and
  Feng}]{41_hou2022vision}
\bibinfo{author}{Q.~Hou}, \bibinfo{author}{Z.~Jiang},
  \bibinfo{author}{L.~Yuan}, \bibinfo{author}{M.~Cheng},
  \bibinfo{author}{S.~Yan}, \bibinfo{author}{J.~Feng},
\newblock \bibinfo{title}{Vision permutator: A permutable mlp-like architecture
  for visual recognition},
\newblock \bibinfo{journal}{IEEE Transactions on Pattern Analysis and Machine
  Intelligence} \bibinfo{volume}{45} (\bibinfo{year}{2022})
  \bibinfo{pages}{1328--1334}.
%Type = Inproceedings
\bibitem[{Tang et~al.(2022)Tang, Zhao, Wang, Luo, Xie, and
  Zeng}]{42_tang2022sparse}
\bibinfo{author}{C.~Tang}, \bibinfo{author}{Y.~Zhao},
  \bibinfo{author}{G.~Wang}, \bibinfo{author}{C.~Luo},
  \bibinfo{author}{W.~Xie}, \bibinfo{author}{W.~Zeng},
\newblock \bibinfo{title}{Sparse mlp for image recognition: Is self-attention
  really necessary?},
\newblock in: \bibinfo{booktitle}{Proceedings of the AAAI Conference on
  Artificial Intelligence}, \bibinfo{year}{2022}, pp.
  \bibinfo{pages}{2344--2351}.
%Type = Article
\bibitem[{Hugo et~al.(2022)Hugo, Piotr, Mathilde, Matthieu, Alaaeldin, Edouard,
  Gautier, Armand, Gabriel, Jakob, and Hervé}]{43_touvron2022resmlp}
\bibinfo{author}{T.~Hugo}, \bibinfo{author}{B.~Piotr},
  \bibinfo{author}{C.~Mathilde}, \bibinfo{author}{C.~Matthieu},
  \bibinfo{author}{E.-N. Alaaeldin}, \bibinfo{author}{G.~Edouard},
  \bibinfo{author}{I.~Gautier}, \bibinfo{author}{J.~Armand},
  \bibinfo{author}{S.~Gabriel}, \bibinfo{author}{V.~Jakob},
  \bibinfo{author}{J.~Hervé},
\newblock \bibinfo{title}{Resmlp: Feedforward networks for image classification
  with data-efficient training},
\newblock \bibinfo{journal}{IEEE Transactions on Pattern Analysis and Machine
  Intelligence} \bibinfo{volume}{45} (\bibinfo{year}{2022})
  \bibinfo{pages}{5314--5321}.
%Type = Inproceedings
\bibitem[{Liu et~al.(2021)Liu, Dai, David, and V}]{44_NEURIPS2021_4cc05b35}
\bibinfo{author}{H.~Liu}, \bibinfo{author}{Z.~Dai}, \bibinfo{author}{S.~David},
  \bibinfo{author}{L.~Q. V},
\newblock \bibinfo{title}{Pay attention to mlps},
\newblock in: \bibinfo{editor}{M.~Ranzato}, \bibinfo{editor}{A.~Beygelzimer},
  \bibinfo{editor}{Y.~Dauphin}, \bibinfo{editor}{P.~Liang},
  \bibinfo{editor}{J.~W. Vaughan} (Eds.), \bibinfo{booktitle}{Advances in
  Neural Information Processing Systems}, \bibinfo{year}{2021}, pp.
  \bibinfo{pages}{9204--9215}.
%Type = Inproceedings
\bibitem[{Timur et~al.(2017)Timur, Alexandre, Franois, Pascal, and
  Silvio}]{45_bagautdinov2017social}
\bibinfo{author}{B.~Timur}, \bibinfo{author}{A.~Alexandre},
  \bibinfo{author}{F.~Franois}, \bibinfo{author}{F.~Pascal},
  \bibinfo{author}{S.~Silvio},
\newblock \bibinfo{title}{Social scene understanding: End-to-end multi-person
  action localization and collective activity recognition},
\newblock in: \bibinfo{booktitle}{Proceedings of the IEEE conference on
  computer vision and pattern recognition}, \bibinfo{year}{2017}, pp.
  \bibinfo{pages}{4315--4324}.
%Type = Inproceedings
\bibitem[{Christian et~al.(2016)Christian, Vincent, Sergey, Jon, and
  Zbigniew}]{46_szegedy2016rethinking}
\bibinfo{author}{S.~Christian}, \bibinfo{author}{V.~Vincent},
  \bibinfo{author}{I.~Sergey}, \bibinfo{author}{S.~Jon},
  \bibinfo{author}{W.~Zbigniew},
\newblock \bibinfo{title}{Rethinking the inception architecture for computer
  vision},
\newblock in: \bibinfo{booktitle}{Proceedings of the IEEE conference on
  computer vision and pattern recognition}, \bibinfo{year}{2016}, pp.
  \bibinfo{pages}{2818--2826}.
%Type = Inproceedings
\bibitem[{He et~al.(2017)He, Georgia, Piotr, and Ross}]{47_he2017mask}
\bibinfo{author}{K.~He}, \bibinfo{author}{G.~Georgia},
  \bibinfo{author}{D.~Piotr}, \bibinfo{author}{G.~Ross},
\newblock \bibinfo{title}{Mask r-cnn},
\newblock in: \bibinfo{booktitle}{Proceedings of the IEEE international
  conference on computer vision}, \bibinfo{year}{2017}, pp.
  \bibinfo{pages}{2961--2969}.
%Type = Inproceedings
\bibitem[{Yuan and Ni(2021)}]{48_yuan2021learning}
\bibinfo{author}{H.~Yuan}, \bibinfo{author}{D.~Ni},
\newblock \bibinfo{title}{Learning visual context for group activity
  recognition},
\newblock in: \bibinfo{booktitle}{Proceedings of the AAAI Conference on
  Artificial Intelligence}, \bibinfo{year}{2021}, pp.
  \bibinfo{pages}{3261--3269}.
%Type = Inproceedings
\bibitem[{Hu et~al.(2020)Hu, Cui, He, and Yu}]{50_hu2020progressive}
\bibinfo{author}{G.~Hu}, \bibinfo{author}{B.~Cui}, \bibinfo{author}{Y.~He},
  \bibinfo{author}{S.~Yu},
\newblock \bibinfo{title}{Progressive relation learning for group activity
  recognition},
\newblock in: \bibinfo{booktitle}{Proceedings of the IEEE/CVF Conference on
  Computer Vision and Pattern Recognition}, \bibinfo{year}{2020}, pp.
  \bibinfo{pages}{980--989}.
%Type = Article
\bibitem[{Yan et~al.(2021)Yan, Shu, Yuan, Tian, and Tang}]{51_yan2021position}
\bibinfo{author}{R.~Yan}, \bibinfo{author}{X.~Shu}, \bibinfo{author}{C.~Yuan},
  \bibinfo{author}{Q.~Tian}, \bibinfo{author}{J.~Tang},
\newblock \bibinfo{title}{Position-aware participation-contributed temporal
  dynamic model for group activity recognition},
\newblock \bibinfo{journal}{IEEE Transactions on Neural Networks and Learning
  Systems} \bibinfo{volume}{33} (\bibinfo{year}{2021})
  \bibinfo{pages}{7574--7588}.
%Type = Inproceedings
\bibitem[{Wang et~al.(2017)Wang, Ni, and Yang}]{52_wang2017recurrent}
\bibinfo{author}{M.~Wang}, \bibinfo{author}{B.~Ni}, \bibinfo{author}{X.~Yang},
\newblock \bibinfo{title}{Recurrent modeling of interaction context for
  collective activity recognition},
\newblock in: \bibinfo{booktitle}{Proceedings of the IEEE conference on
  computer vision and pattern recognition}, \bibinfo{year}{2017}, pp.
  \bibinfo{pages}{3048--3056}.
%Type = Article
\bibitem[{der Maaten~Laurens and Geoffrey(2008)}]{49_van2008visualizing}
\bibinfo{author}{V.~der Maaten~Laurens}, \bibinfo{author}{H.~Geoffrey},
\newblock \bibinfo{title}{Visualizing data using t-sne.},
\newblock \bibinfo{journal}{Journal of machine learning research}
  \bibinfo{volume}{9} (\bibinfo{year}{2008}) \bibinfo{pages}{2579--2605}.

\end{thebibliography}
\end{document}